%% file: main.tex
\documentclass{article}
\usepackage{spconf,amsmath,graphicx}
\usepackage{subcaption}
\usepackage[switch]{lineno}
\usepackage{cite}
\usepackage{amssymb,amsfonts}
\usepackage{bm} 

\usepackage{algorithm, algorithmic}
\usepackage{textcomp}
\usepackage{xcolor}
\usepackage{booktabs}
\usepackage{multirow}
\usepackage{makecell}
\usepackage{url}
\usepackage{upgreek}

\usepackage{kotex} 
\usepackage{pgfplots}
\pgfplotsset{compat=1.17}

\usepackage{color}

\input{data/framewise_results}

\usepackage{fancyhdr}
\usepackage{hyperref}

\usepackage{textcomp}
\usepackage[absolute]{textpos}

\newcommand{\copyrightstatement}{
    \begin{textblock}{13}(1.3,0.2)
    \noindent
    \textblockcolour{white}
    \centering
    \parbox{\textwidth}{
        \centering
        \footnotesize
        \copyright\ 2025 IEEE.  Personal use of this material is permitted. Permission from IEEE must be obtained for all other uses, in any current or future media, including reprinting/republishing this material for advertising or promotional purposes, creating new collective works, for resale or redistribution to servers or lists, or reuse of any copyrighted component of this work in other works.
    }
    \end{textblock}
}

\begin{document}

\thispagestyle{empty}
\copyrightstatement

\title{Autoregression-free video prediction using diffusion model for mitigating error propagation}

\name{%
    Woonho Ko$^1$
    \qquad Jin Bok Park$^1$%
    \qquad Il Yong Chun$^{1,2,3, \dagger}$%
    \thanks{
    $^\dag$Corresponding author.
    Emails of authors:
    \noindent 
    $\{ \textsf{heesuk988@g.skku.edu}$, 
    $\textsf{bjb663@g.skku.edu}$,
    $\textsf{iychun@skku.edu} \}$.
    The work of W. H. Ko, J. B. Park and I. Y. Chun was supported in part by 
    NRF Grant RS-2023-00213455 funded by MSIT,
    the 2024 Digital Therapeutics Development and Demonstration Support Program funded by MSIT and NIPA,
    and the BK21 FOUR Project. 
    The work of I. Y. Chun was additionally supported in part by 
    IITP Grant RS-2019-II190421 funded by MSIT, 
    KIAT Grant RS-2024-00418086 funded by MOTIE,
    and IBS-R015-D1.
    }
}

\address{%
   $^1$Dept.~of Electrical \& Computer Engn., Sungkyunkwan University (SKKU), Suwon 16419, South Korea \\%
   $^2$Depts.~of Artificial Intelligence, Advanced Display Engn., and Semiconductor Converg.~Engn., SKKU \\%
   $^3$Center for Neuroscience Imaging Research, Inst.~for Basic Science (IBS), Suwon 16419, South Korea
}
\maketitle
\begin{abstract}
\vspace{-0.1pc}
Existing long-term video prediction methods often rely on an autoregressive video prediction mechanism.
However, this approach suffers from error propagation, particularly in distant future frames.
To address this limitation, this paper proposes the first AutoRegression-Free (ARFree) video prediction framework using diffusion models.
Different from an autoregressive video prediction mechanism, ARFree directly predicts any future frame tuples from the context frame tuple.
The proposed ARFree consists of two key components:
\textit{1)}
a motion prediction module that predicts a future motion using motion feature extracted from the context frame tuple;
\textit{2)}
a training method that improves motion continuity and contextual consistency between adjacent future frame tuples.
Our experiments with two benchmark datasets show that the proposed ARFree video prediction framework outperforms several state-of-the-art video prediction methods.
Our codes are available at: \href{https://github.com/kowoonho/ARFree}{\url{https://github.com/kowoonho/ARFree}}.

\end{abstract}
\vspace{-0.7pc}
\section{Introduction}
\vspace{-0.7pc}
\label{sec:intro}
Video prediction is a challenging computer vision task.
It aims to predict plausible future frames from past frames.
By predicting future frames at the pixel level, 
video prediction can be used in various downstream tasks, such as decision-making systems \cite{decision-making}, autonomous driving \cite{autonomous-driving}, and robotic navigation \cite{robot-navigation}. 
However, predicting future frames remains extremely challenging due to the inherent uncertainty in future events. 
This uncertainty grows exponentially in long-term video predictions.

Existing long-term video prediction methods rely on an autoregressive (AR) video prediction mechanism that uses previously predicted frames for subsequent predictions of future frames.
However, this mechanism suffers from error propagation.
Prediction errors from previously predicted frames go into subsequent predictions of future frames, leading to significant performance degradations, particularly in predicting distant future frames.

\begin{figure*}[t]
    \centering
    \centerline{\includegraphics[width=0.8\textwidth]{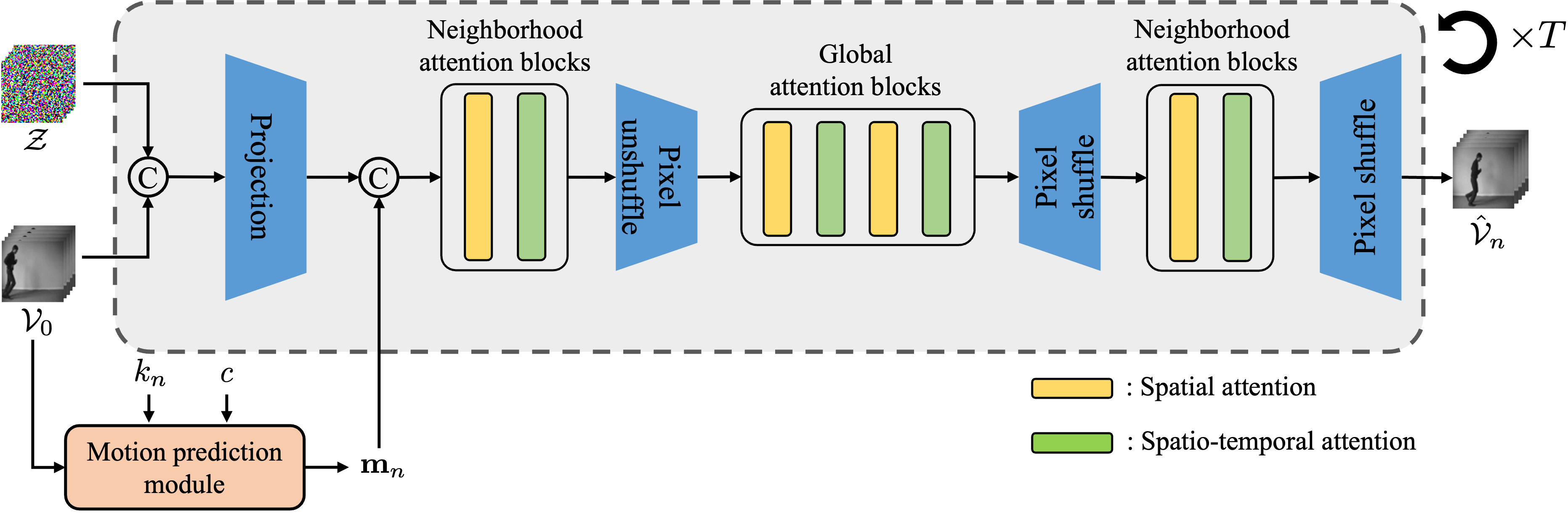}}
    \vspace{-0.7pc}
    \caption{
    The overall architecture of the proposed ARFree video prediction diffusion model in the reverse diffusion process.
    The model predicts the $n$th future frame tuple by iteratively denoising pure noise frame from the $\cZ$ over $T$ steps, given its corresponding motion feature $\mb{m}_n$ and the context frame tuple $\cV_0$. 
    We use the ARFree motion prediction module that extracts $\mb{m}_n$ from $\cV_0$ using the frame index of the $n$th future frame tuple, $k_n$, and the corresponding class label $c$.
    }
    \vspace{-1.2pc}
    \label{fig:architecture}
\end{figure*}

To address this limitation, 
we propose the first AutoRegre-ssion-Free (ARFree) video prediction framework using diffusion models.
Different from existing AR video prediction mechanism, 
the proposed ARFree video prediction mechanism 
eliminates the dependency on previously predicted frames to mitigate the error propagation issue.
This mechanism uses the context frame tuple $\cV_0$ (i.e., past frame tuple) for predicting any future frame tuples $\cV_n$ for $n=1,\ldots,N$, where $N$ denotes the total number of future frame tuples.

The ARFree video prediction mechanism has two challenges.
The first challenge is to model temporal relation between the context frame tuple $\cV_0$ and a future frame tuple $\cV_n$, $n \in \{ 1,\ldots,N \}$.
In the ARFree video prediction mechanism, 
there could be a large temporal discontinuity between $\cV_0$ and $\cV_n$, particularly for $n \in \{2,\ldots,N\}$,
making it difficult to model their temporal relation.
The second challenge is to improve motion continuity and contextual consistency between adjacent future frame tuples.
Particularly in diffusion-based video prediction models, 
the stochastic nature of the diffusion process induces randomness in predictions. 
Without appropriate constraint, this randomness can lead to motion discontinuities and contextual inconsistencies between adjacent future frame tuples, 
degrading the overall naturalness of the generated video sequence. 

This paper proposes two comprehensive solutions to overcome the two aforementioned challenges.
First, to explicitly model the temporal relation between the context frame tuple and a future frame tuple, 
we propose an ARFree motion prediction module that predicts future motion information from the context frame tuple.
We then incorporate the predicted future motion information as a conditioning input in the proposed ARFree diffusion model, improving its prediction capability for long-term motion patterns.
Second, to improve the motion continuity and the contextual consistency between adjacent future frame tuples,
we propose a training method that uses frames in an overlap between time windows from adjacent future frame tuples.
Our contributions are summarized as follows:
\bulls{
    \item We propose the first \textbf{ARFree} video prediction framework that can moderate the critical error propagation issue in existing AR video prediction approaches.

    \item We propose an ARFree motion prediction module to explicitly model the temporal relation between the context frame tuple and a future frame tuple.
    
    \item We propose a training method to improve the motion continuity and the contextual consistency between adjacent future frame tuples.
    
    \item Our experiments with the KTH~\cite{kth} and NATOPS datasets~\cite{natops} show that the proposed ARFree video prediction framework outperforms several existing state-of-the-art (SOTA) video prediction methods.
}

\vspace{-0.7pc}
\section{Related works}
\label{sec:related works}
\vspace{-0.7pc}
This section reviews existing video prediction diffusion models.
With the emergence of diffusion models \cite{ddpm}, there have been efforts to improve prediction quality in video prediction using diffusion models. 
Residual Video Diﬀusion (RVD) predicts residual error of the next video frame using diffusion models \cite{RVD}.
Masked Conditional Video Diffusion (MCVD) proposed a unified framework that predicts masked frames using diffusion models for video prediction, generation, and interpolation \cite{mcvd}.

However, these methods directly use past frames as input, without predicting motion information.
Different from the above methods,
Spatio-Temporal Diffusion (STDiff) extracts motion information from pixel-wise differences between two adjacent frames in a tuple of past frames and predicts future motion \cite{stdiff} using a neural stochastic differential equation \cite{neural-sde}. 
This predicted motion is then incorporated as a conditioning input in diffusion models.
Latent Flow Diffusion Model (LFDM) \cite{lfdm} and Distribution Extrapolation Diffusion Model (ExtDM) \cite{extdm} further enrich the motion information by predicting future motion through a diffusion model and decoding it to predict future frames.
Yet, the error propagation issue in long-term video prediction remains unresolved.

\vspace{-0.7pc}
\section{Backgrounds}
\label{sec:background}
\vspace{-0.7pc}
This section includes some backgrounds of diffusion models.
Diffusion models \cite{ddpm} are defined by the forward and reverse processes.
The forward process iteratively adds isotropic Gaussian noise to a clean sample $\mb{x}_0$:  
$\mb{x}_{\sigma} = \mb{x}_0 + \sigma \boldsymbol{\upepsilon}$,
where $\boldsymbol{\upepsilon} \sim \cN(\mb{0}, \mb{I})$.
The standard deviation $\sigma$ of the Gaussian noise is predefined by a monotonically increasing noise level schedule: $\sigma_\text{min} \leq \sigma \leq \sigma_\text{max}$, where $\sigma_\text{min}$ and $\sigma_\text{max}$ denote the predefined minimum and maximum value of $\sigma$, respectively.
The reverse process starts with a pure noise sample 
$\mb{x}_{\sigma_\text{max}} \sim \cN(\mb{0}, \sigma_\text{max}^2 \mb{I})$ and gradually denoises it using a trained denoiser 
$D_{\boldsymbol{\uptheta}}(\mb{x}_{\sigma}, \sigma)$ with parameters ${\boldsymbol{\uptheta}}$, 
generating a clean sample $\mb{x}_0$.

We follow the EDM formulation to optimize the denoiser $D_{\boldsymbol{\uptheta}}$ using the following objective \cite{edm}:
\be{
    \label{eq:diffusion}
    \mathbb{E}_{\mb{x}_0,\boldsymbol{\upepsilon}, \sigma}
    \left[
    \lambda_{\sigma} \| D_{\boldsymbol{\uptheta}}(\mb{x}_{\sigma}, \sigma) - \mb{x}_0 \|_2^2
    \right],
}
where $\lambda_{\sigma}$ is a weighting function depending on $\sigma$. 
To train the denoiser, 
EDM uses the preconditioning scheme to scale both the input and output of the objective function.
The denoiser is defined as follows \cite{edm}:
\eas{
    D_{\boldsymbol{{\uptheta}}}(\mb{x}_{\sigma}, \sigma) 
    =
    c_{\text{out}}(\sigma) 
    D'_{\boldsymbol{\uptheta}} \big( 
        c_{\text{in}}(\sigma) \mb{x}_{\sigma}, 
        c_{\text{noise}}(\sigma) 
    \big) \nonumber + c_{\text{skip}}(\sigma) \mb{x}_{\sigma},
}
where $D'_{\boldsymbol{\uptheta}}$ is a denoising neural network to be trained, 
and $c_{\text{in}}(\sigma)$, $c_{\text{out}}(\sigma)$, $c_{\text{noise}}(\sigma)$, and $c_{\text{skip}}(\sigma)$ are predefined preconditioning terms that modulate the input and output at different noise levels.
\vspace{-0.7pc}
\section{Methodology}
\label{sec:method}
\vspace{-0.7pc}
This section proposes the first ARFree video prediction diffusion framework.
Fig.~\ref{fig:architecture} illustrates its overall architecture.
The proposed ARFree video prediction diffusion model predicts 
the $n$th future frame tuple $\cV_n$ for $n=1, \ldots, N$, 
given two conditions, \textit{1)} its corresponding motion feature $\mb{m}_n$ and \textit{2)} the context frame tuple $\cV_0$.
We define $\cV_n$ and $\cV_0$ as follows:
\ea{
& \cV_{n} := \left( \mb{v}_i : i = F_\text{p} + (n-1)F_\text{f}, \ldots, F_\text{p} + n F_\text{f} - 1 \right), 
\label{eq:vn}
\\
& \cV_{0} := \left( \mb{v}_i : i = 0,\ldots, F_\text{p} - 1 \right),
\nonumber \label{eq:v0}
}
where $\mb{v}_i$ denotes the $i$th video frame in a video sequence, and $F_\text{f}$ and $F_\text{p}$ denote the number of frames in the future and context frame tuples, respectively.
We iteratively denoise pure noise frames from the tuple $\cZ := (\mb{z}_i \sim \cN(\mb{0}, \sigma_\text{max}^2 \mb{I}): i=1,\ldots,F_\text{f})$ over $T$ steps using the proposed ARFree video prediction diffusion model, resulting in the predicted $n$th future frame tuple $\hat{\cV}_n$.

To explicitly model the temporal relation between the context frame tuple and a future frame tuple (see Section~\ref{sec:intro}),
we propose a motion prediction module that gives $\mb{m}_{n}$ from $\cV_{0}$, 
using the initial frame index of the $n$th future frame tuple $k_n := F_\text{p} + (n-1) F_\text{f}$, and the corresponding class label $c \in \{ 1, \ldots, C \}$, $\forall n$.
For a denoising neural network in the ARFree video prediction diffusion model, we modify the Hourglass Diffusion Transformer (HDiT) architecture \cite{hdit}.
Sections~\ref{sec: motion prediction} and \ref{sec:hdit} describe the proposed motion prediction module and denoising neural network architecture, respectively.
To improve the motion continuity and the contextual consistency between adjacent future frame tuples (see Section~\ref{sec:intro}),
Section~\ref{sec: training method} proposes a training method.

\begin{figure}[t]
    \centering
    \centerline{\includegraphics[width=\linewidth]{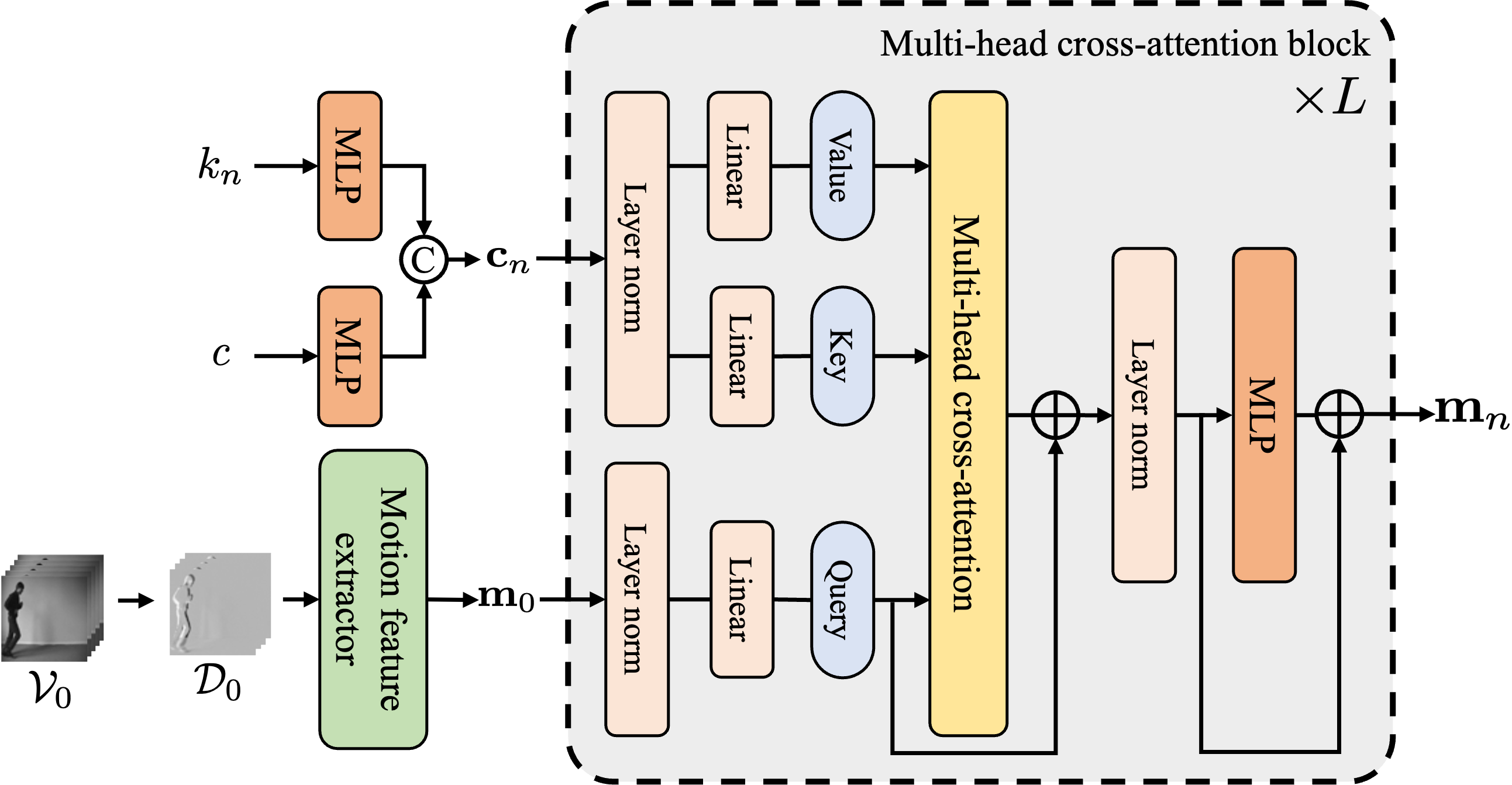}}
    \vspace{-0.6pc}
    \caption{
    The architecture of the proposed motion prediction module.
    For query, we extract the motion feature for $\cV_0$, $\mb{m}_0$ in (\ref{eq:conv-gru}).
    For key and value, we use transformed $\mb{c}_n$'s.
    We pass query, key, and value through $L$ multi-head cross-attention blocks and ultimately, predict the motion feature corresponding to $\cV_n$, $\mb{m}_n$ in (\ref{eq:motion predictor}).
    }
    \vspace{-1.2pc}
    \label{fig:motion prediction module}
\end{figure}

\vspace{-0.7pc}
\subsection{Proposed ARFree motion prediction module}
\label{sec: motion prediction}
\vspace{-0.5pc}
To extract the motion feature for $\cV_0$,
$\mb{m}_0$,
we use the gated recurrent unit-based motion feature extractor $E_{\boldsymbol{\uptheta}_\text{e}}$ having parameters $\boldsymbol{\uptheta}_\text{e}$ \cite{mcnet}:
\be{
    \label{eq:conv-gru}
    \mb{m}_0 = E_{\boldsymbol{\uptheta}_\text{e}}(\cD_0),
}
where $\cD_0 = \left( \mb{v}_i - \mb{v}_{i-1} : i=1, \ldots, F_\text{p}-1 \right)$ denotes a tuple of pixel-wise differences between two adjacent frames in $\cV_0$.

Now, we propose a motion prediction module $P_{\boldsymbol{\uptheta}_\text{p}}$ with parameters $\boldsymbol{\uptheta}_\text{p}$ that predicts the future motion feature for $\cV_n$, $\mb{m}_{n}$.
We pass $k_n$ and $c$ through its each two-layer multi-layer perceptron network and concatenate their output to form $\mb{c}_{n}$.
The motion prediction module $P_{\boldsymbol{\uptheta}_\text{p}}$ consists of $L$ multi-head cross-attention blocks.
In each cross-attention block, 
we embed $\mb{m}_0$ obtained in (\ref{eq:conv-gru}) as a query.
To incorporate the time and action information, 
we embed $\mb{c}_n$ obtained above as key and value for each cross-attention block.
By passing the above query, key, and value through multi-head cross-attention blocks in the $P_{\boldsymbol{\uptheta}_\text{p}}$, 
we finally obtain the future motion feature corresponding to the $n$th future frame tuple, $\mb{m}_n$:
\be{
    \label{eq:motion predictor}
    \mb{m}_n = P_{\boldsymbol{\uptheta}_\text{p}}(\mb{m}_0, k_n, c).
}
Fig.~\ref{fig:motion prediction module} illustrates the architecture of the motion prediction module.
By explicitly modeling future motion feature, 
the proposed module in (\ref{eq:motion predictor}) models the temporal relation between the context frame tuple $\cV_0$ and the $n$th future frame tuple $\cV_n$. 

As a conditioning method in the proposed  diffusion model,
we simply concatenate its input with $\mb{m}_{n}$ in (\ref{eq:motion predictor}).

\vspace{-0.7pc}
\subsection{Modified HDiT architecture for video prediction}
\label{sec:hdit}
\vspace{-0.5pc}
This section modifies the HDiT architecture~\cite{hdit} for the proposed ARFree video prediction diffusion models.
The original HDiT architecture, designed for image generation~\cite{hdit}, contains a series of neighborhood attention blocks and global attention blocks with pixel unshuffle and shuffle \cite{unshuffle}.
Each attention block is constructed with spatial attention layers.
The pixel unshuffle and shuffle are for downsampling and upsampling, respectively.

For the proposed ARFree video prediction framework, 
we add spatio-temporal attention layers after every spatial attention layers, inspired by \cite{MVDream}.
In each spatio-temporal attention layer,
we reshape a video tensor by combining its spatial and temporal dimension into a single sequence of tokens.
By integrating spatio-temporal information,
this attention mechanism promotes temporal correlations between a past frame and a distant future frame, different from the one-dimensional temporal attention mechanism commonly used in existing video diffusion models \cite{vdm, lvdm}.
To incorporate frame index information for a future frame tuple,
we apply rotary position embedding to each attention layer \cite{rope}.

\begin{figure}[t!]
    \centering
    \centerline{\includegraphics[width=0.9\linewidth]{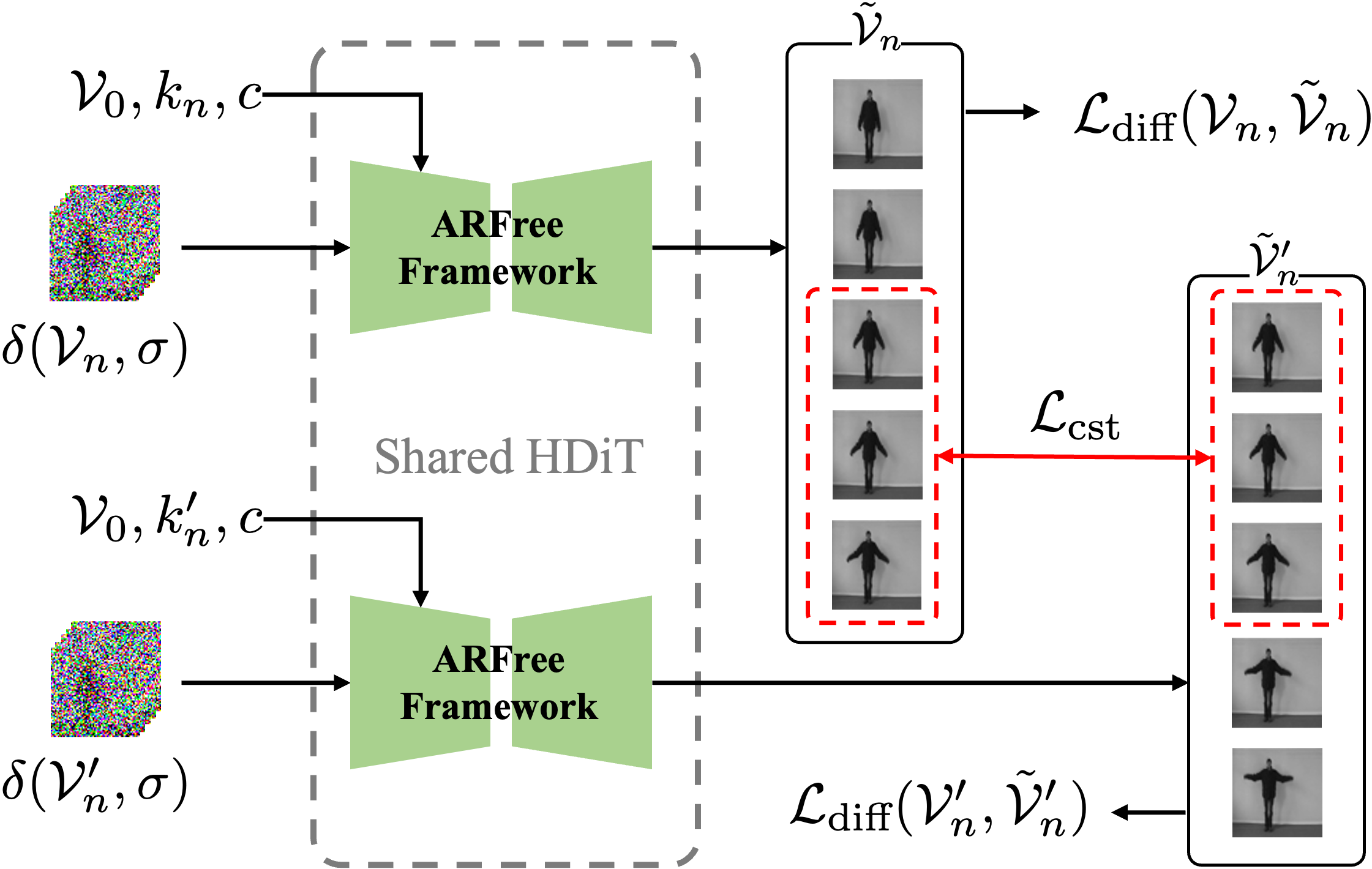}}
    \vspace{-0.6pc}
    \caption{
    The proposed training pipeline of the ARFree video prediction diffusion model framework.
    In training the proposed diffusion model, we denoise two noisy future frame tuples with overlapping time windows.
    To improve the motion continuity and the contextual consistency between two adjacent future frame tuples, we use two types of learning objectives:  $\cL_{\text{diff}}$ and $\cL_\text{cst}$ in (\ref{eq:diffusion loss}) and (\ref{eq:consistency loss}), respectively.
    }
    \label{fig:training}
    \vspace{-1.2pc}
\end{figure}

\vspace{-0.7pc}
\subsection{Proposed training method}
\label{sec: training method}
\vspace{-0.5pc}
This section proposes a new training loss that can improve the motion continuity and the contextual consistency between two adjacent future frame tuples (see Section~\ref{sec:intro}).
In the proposed training method, 
the ARFree diffusion model $D_{\boldsymbol{\uptheta}}$ with parameters $\boldsymbol{\uptheta}$ denoises two noisy future frame tuples with overlapping time windows, $\delta(\cV_n, \sigma)$ and $\delta(\cV'_n, \sigma)$,
where $\delta(\cA, \sigma)$ denotes the forward process that adds isotropic Gaussian noise predefined by noise level schedule $\sigma$ to frames in tuple $\cA$.
The future frame tuple $\cV'_n$ includes at least one frame in $\cV_n$ (\ref{eq:vn}):
\bes{
\cV'_n := ( \mb{v}_i : i=F_\text{p} + (n-1)F_\text{f} + F_\text{o}, \ldots, F_\text{p} + nF_\text{f} - 1 + F_\text{o} ),
}
for $F_\text{o} \in \{ 1, \ldots, F_\text{f} - 1 \}$, 
where $F_\text{o}$ denotes the number of frames in an overlap between time windows from $\cV_n$ and $\cV'_n$.
We write denoised versions of $\delta(\cV_n, \sigma)$ and $\delta(\cV'_n, \sigma)$, respectively, as follows:
\eas{
& \tilde{\cV}_n = D_{{\boldsymbol{\uptheta}}}(\delta(\cV_n, \sigma), \cV_0, k_n, c, \sigma), \\
& \tilde{\cV}'_n = D_{{\boldsymbol{\uptheta}}}(\delta(\cV'_n, \sigma), \cV_0, k'_n, c, \sigma),
}
where $k'_n := F_\text{p} + (n-1) F_\text{f} + F_\text{o}$ denotes the initial frame index of $\cV'_n$.

First, we apply the EDM loss function(\ref{eq:diffusion}) to the $n$th future frame tuple $\cV_n$ (\ref{eq:vn}):
\be{
    \label{eq:diffusion loss}
    \cL_{\text{diff}}(\cV_n, \tilde{\cV}_n) = \mathbb{E}_{\boldsymbol{\upepsilon}, \sigma}
    \left[
    {\lambda_{\sigma} \over F_\text{f}}
    \sum_{i=0}^{F_\text{f}-1}
    \left\| \cV_{n}(i) - \tilde{\cV}_{n}(i) \right\|_2^2
    \right],
}
In the same manner, 
we apply (\ref{eq:diffusion}) to the future frame tuple $\cV'_n$ that has an overlap time window with $\cV_n$, by defining the loss $\cL_\text{diff}(\cV'_n, \tilde{\cV}'_n)$.
Note that we share the denoiser $D_{{\boldsymbol{\uptheta}}}$ in both losses.

Yet, 
in the aforementioned two losses, $\cL_{\text{diff}}(\cV_n, \tilde{\cV}_n)$ and $\cL_\text{diff}(\cV'_n, \tilde{\cV}'_n)$,
we do not model the motion continuity and contextual consistency
between two adjacent future frame tuples.
We aim to improve the above two properties by promoting the similarity between frames in an overlap between time windows of $\tilde{\cV}_n$ and $\tilde{\cV}'_n$:
\be{
\label{eq:consistency loss}
\cL_\text{cst} := 
\bbE_{\boldsymbol{\upepsilon}, \sigma}
\left[
{\lambda_{\sigma} \over F_\text{o}}
\sum_{i=0}^{F_\text{o}-1}
\left\| \tilde{\cV}_{n,o}(i) - \tilde{\cV}'_{n,o}(i) \right\|_2^2
\right],
}
where $\tilde{\cV}_{n,\text{o}}$ and $\tilde{\cV}'_{n,\text{o}}$ denote tuples of frames in an overlap between time windows from $\tilde{\cV}_n$ and $\tilde{\cV}'_n$, respectively.

Our total training loss $\cL_\text{total}$ is written by:
\bes{
\cL_\text{total} = 
 \mathbb{E}_{\cX}
\left[
{1 \over 2}
\left(
\cL_\text{diff}(\cV_n, \tilde{\cV}_n) + \cL_\text{diff}(\cV'_n, \tilde{\cV}'_n)
\right) + \lambda \cL_\text{cst}
\right],
}
where $\cX := \{\cV_n, \cV'_n, \cV_0, k_n, k'_n, c\}$ denotes the set of inputs in the diffusion model.
We use $\lambda = 0.1$ as the default weight.
Fig.~\ref{fig:training} illustrates the proposed training method.

\begin{figure}[t!]
\centering
\setlength{\tabcolsep}{0.6pt}
\renewcommand{\arraystretch}{0.5}
    \begin{tabular}{c c | c c c c c}
    \multicolumn{2}{c|}{Context frames} & \multicolumn{5}{c}{Predicted frames} \\
    
    \small $t=6$ & \small $t=10$ & \small $t=14$ & \small $t=18$ & \small $t=22$ & \small $t=26$ & $t=30$ \\
    \includegraphics[scale=0.5]{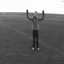} &
    \includegraphics[scale=0.5]{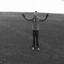} &
    \includegraphics[scale=0.5]{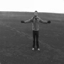} &
    \includegraphics[scale=0.5]{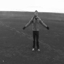} &
    \includegraphics[scale=0.5]{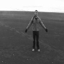} &
    \includegraphics[scale=0.5]{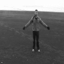} &
    \includegraphics[scale=0.5]{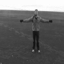}
    \\
    \multicolumn{2}{c|}{\raisebox{2.2\height}{\textbf{\small Proposed ARFree}}} &
    \includegraphics[scale=0.5]{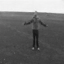} &
    \includegraphics[scale=0.5]{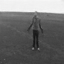} &
    \includegraphics[scale=0.5]{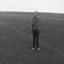} &
    \includegraphics[scale=0.5]{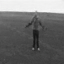} &
    \includegraphics[scale=0.5]{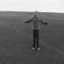} 
    \\
    \multicolumn{2}{c|}{\raisebox{2.2\height}{\small MCVD}} &
    \includegraphics[scale=0.5]{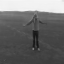} &
    \includegraphics[scale=0.5]{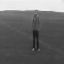} &
    \includegraphics[scale=0.5]{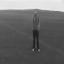} &
    \includegraphics[scale=0.5]{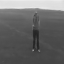} &
    \includegraphics[scale=0.5]{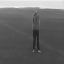}
    \\
    \multicolumn{2}{c|}{\raisebox{2.2\height}{\small STDiff}} &
    \includegraphics[scale=0.5]{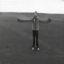} &
    \includegraphics[scale=0.5]{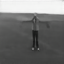} &
    \includegraphics[scale=0.5]{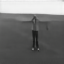} &
    \includegraphics[scale=0.5]{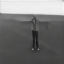} &
    \includegraphics[scale=0.5]{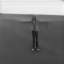}
    \end{tabular}
    \vspace{-0.6pc}
    \caption{
    Qualitative comparisons between different video prediction models (KTH dataset).
    }
    \vspace{-1.2pc}
    \label{fig:qualitative results}
\end{figure}

\begin{table*}[t!]
\centering
\footnotesize
\caption{Comparisons between different video prediction methods with two benchmark datasets}
\vspace{-3mm}
\label{tab:wide_comparison}
\setlength{\tabcolsep}{2pt}
\renewcommand{\arraystretch}{1.0}
\begin{tabular}{c|>{\centering\arraybackslash}p{1.0cm}|>{\centering\arraybackslash}p{1.0cm}|>{\centering\arraybackslash}p{1.0cm}|>{\centering\arraybackslash}p{1.0cm}|>{\centering\arraybackslash}p{1.0cm}|>{\centering\arraybackslash}p{1.0cm}|>{\centering\arraybackslash}p{1.0cm}|>{\centering\arraybackslash}p{1.0cm}|>
{\centering\arraybackslash}p{0.8cm}||>{\centering\arraybackslash}p{1.0cm}|>{\centering\arraybackslash}p{1.0cm}|>{\centering\arraybackslash}p{1.0cm}|>
{\centering\arraybackslash}p{1.0cm}|>
{\centering\arraybackslash}p{0.8cm}}
\hline\hline
\multirow{2}{*}{Methods} & \multicolumn{4}{c|}{KTH ($F_\text{p} = 10 \rightarrow F_\text{total} = 20$)} & \multicolumn{4}{c|}{KTH ($F_\text{p} = 10 \rightarrow F_\text{total} = 30$)} & \multirow{2}{*}{FPS ↑} & \multicolumn{4}{c|}{NATOPS ($F_\text{p} = 10 \rightarrow F_\text{total} = 30$)} & \multirow{2}{*}{FPS ↑}  \\ \cline{2-9} \cline{11-14}
                        & PSNR ↑ & SSIM ↑ & LPIPS ↓ & FVD ↓ & PSNR ↑ & SSIM ↑ & LPIPS ↓ & FVD ↓ &
                        & PSNR ↑ & SSIM ↑ & LPIPS ↓ & FVD ↓ \\ \hline \hline
ExtDM  \cite{extdm}     & 21.44 & 0.678 & 0.217 & 576.4 
                        & 22.16 & 0.711 & 0.167 & 507.8 & \textbf{48.19}
                        & 23.07 & 0.776 & 0.131 & 1345.9  & \textbf{10.95}     \\ \hline
LFDM \cite{lfdm}        & 22.87 & 0.699 & 0.138 & 343.4 
                        & 22.47 & 0.685 & 0.154 & 467.6 & 2.03
                        & 25.24 & 0.853 & 0.085 & 473.1 & 1.09   \\ \hline
RVD    \cite{RVD}       & 25.47 & 0.810 & 0.097 & 278.5
                        & 24.18 & 0.769 & 0.125 & 434.7 & 0.10
                        & 23.12 & 0.797 & 0.161 & 4027.1& 0.09     \\ \hline
STDiff \cite{stdiff}    & 25.31 & 0.807 & 0.082 & 179.4 
                        & 24.16 & 0.773 & 0.107 & 282.6 & 0.76
                        & 9.09  & 0.618 & 0.318 & 990.8 & 0.61     \\ \hline
MCVD   \cite{mcvd}      & 26.07 & \textbf{0.850} & 0.083 & 148.2 
                        & 25.04 & \textbf{0.823} & 0.102 & 223.2 & 2.47
                        & 14.97 & 0.690          & 0.326 & 1443.6 & 1.05 
                        \\ \hline

\textbf{ARFree (Ours)}  & \textbf{26.83} & 0.844 & \textbf{0.055} & \textbf{120.0} 
                        & \textbf{25.74} & 0.813 & \textbf{0.068} & \textbf{197.1} & 4.41
                        & \textbf{28.49} & \textbf{0.913} & \textbf{0.054} & \textbf{339.2} & 0.86\\ \hline
\end{tabular}
\vspace{-1.3pc}
\end{table*}

\begin{table}[t!]
    \caption{Comparisons between different ARFree variants with the KTH dataset ($F_\text{total} = 30$)}
    \vspace{-3mm}
    \label{table:ablation}
    \centering
    \begin{tabular}{c c | c c c c}
    \toprule
        {A} & {B} & {FVD ↓} & {PSNR ↑} & {SSIM ↑} & {LPIPS ↓} \\
    \midrule
        X & X & 226.7 & 25.25 & 0.796 & 0.076 \\
        O & X & 210.8 & 25.54 & 0.802 & 0.074 \\
        X & O & 220.4 & 25.53 & 0.806 & 0.072 \\
        O & O & \textbf{197.1} & \textbf{25.74} & \textbf{0.813} & \textbf{0.068} \\
    \bottomrule
    \end{tabular}
    \vspace{-1.0pc}
\end{table}
\vspace{-0.5pc}

\section{Results and discussion}
\label{sec: experiments}
\vspace{-0.7pc}
This section describes the experimental setups and presents results with some discussion. We compared the proposed ARFree video prediction framework with several SOTA video prediction methods.
In addition, we investigate the contribution of different ARFree variants.

\vspace{-0.7pc}
\subsection{Datasets and evaluation metrics}
\vspace{-0.5pc}
We ran experiments with two benchmark datasets, the KTH~\cite{kth} and NATOPS~\cite{natops} datasets.
The \textbf{KTH} dataset consists of videos of $25$ people performing $C=6$ types of actions: \textit{running}, \textit{jogging}, \textit{walking}, \textit{boxing}, \textit{hand-clapping}, and \textit{hand-waving}.
For each video, we resized the grayscale video frames to $64 \!\times\! 64$.
We used the videos of $20$ people for training and five people for test.
The \textbf{NATOPS} dataset includes videos of $20$ people performing $C=24$ types of body-and-hand gestures used for communicating with the U.S. Navy pilots.
We changed the spatial resolution of input color video frames to $128 \!\times\! 128$.
We used videos of ten people for training and another ten people for test.

For training, we randomly sampled $120,\!000$ samples from the videos. 
In each sample, we used \textit{1)} a context frame tuple $\cV_0$, \textit{2)} two adjacent future frame tuples $\cV_n$ and $\cV'_n$, \textit{3)} their initial frame indexes $k_n$ and $k'_n$, and \textit{4)} the corresponding class label $c$.
For test, we randomly sampled $128$ samples from the videos.
In each sample, we used \textit{1)} a context frame tuple $\cV_0$, \textit{2)} a future video sequence of $F_\text{total}$ frames, and \textit{3)} the corresponding class label $c$.
We remark that the proposed diffusion model predicts the $n$th future frame tuple $\cV_n$ for $n=1, \ldots, F_\text{total}/F_\text{f}$, using the context frame tuple $\cV_0$ for test.
We combined the predicted future frame tuples to generate the total video sequence of $F_\text{total}$ frames.
We set $F_\text{p}$ and $F_\text{f}$ as $10$ and $5$, respectively.  
We evaluated the proposed ARFree video prediction framework with $F_\text{total}=20$ and $F_\text{total}=30$ for the KTH dataset, and $F_\text{total}=30$ for the NATOPS dataset.

\noindent\textbf{Evaluation metrics.} 
As evaluation metrics,
we used peak signal-to-noise ratio (PSNR), structural similarity index measure (SSIM), learned perceptual image patch similarity (LPIPS), Fréchet video distance (FVD), and frames-per-second (FPS).

\vspace{-0.7pc}
\subsection{Implementation details}
\vspace{-0.5pc}
As the motion prediction module, we set the number of multi-head cross-attention blocks $L$ as $6$.
For the modified HDiT architecture of the ARFree video prediction diffusion model, 
we used two neighborhood attention blocks and ten global attention blocks, 
with the number of channels set to $256$ and $512$, respectively. 
We trained the ARFree video prediction diffusion model for $400$K steps with a batch size of $16$ and used AdamW optimizer with a learning rate of $10^{-4}$.

In training the diffusion model,
we followed the default training setup described in \cite{hdit}, 
and used the exponential moving average of model weights with a decay factor of $0.995$.
For the reverse process of the diffusion model,
we used the linear multistep method with $T=50$ sampling steps \cite{lms}.
To improve the contextual consistency between the video frames, 
we initialized noise for each video frame with the mixed noise model, similar to \cite{pyoco}.
In reproducing existing SOTA methods, we used their default setups specified in the corresponding papers.
We conducted all experiments with an NVIDIA GeForce RTX 4090 GPU.

\vspace{-0.7pc}
\subsection{Comparisons between different video prediction methods}
\vspace{-0.5pc}
Fig.~\ref{fig:qualitative results} and Table~\ref{tab:wide_comparison}
show that the proposed ARFree video prediction framework outperforms four existing SOTA video prediction methods.
Fig.~\ref{fig:qualitative results} shows that the proposed ARFree video prediction framework can achieve more accurate object shapes and improved motion continuities, compared to the existing SOTA methods with the KTH dataset, particularly in distant future frames.
Table~\ref{tab:wide_comparison} shows that the proposed ARFree video prediction framework achieves better video prediction performance compared to other SOTA video prediction methods with the KTH and NATOPS datasets. 

The results in the tenth and fifteenth columns of Table~\ref{tab:wide_comparison} show that 
it is challenging to achieve real-time video prediction with all the diffusion models (including ours) except ExtDM \cite{extdm}.
ExtDM generates low-resolution motion cues rather than full video frames, and this can accelerate the entire video prediction process.

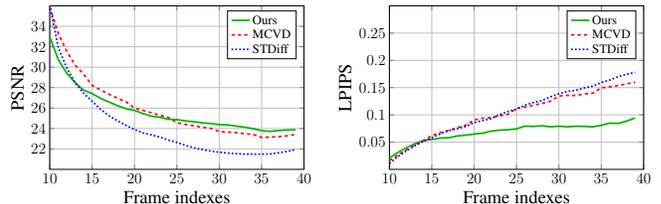
\begin{figure}[t!]
\centering%
\resizebox{\columnwidth}{!}{%
\hspace*{-1mm}%
\begin{tikzpicture}%
\begin{axis}[
  width=100mm, height=70mm,
  xmin={10}, xmax={40}, xtick={10, 15, 20, 25, 30, 35, 40}, xticklabels={$10$, $15$, $20$, $25$, $30$, $35$, $40$},
  ymin={20}, ymax={36}, ytick={22, 24, 26, 28, 30, 32, 34}, yticklabels={$22$, $24$, $26$, $28$, $30$, $32$, $34$},
  grid={major}, legend pos={north east}, legend cell align={left}, legend style={font=\large},
  tick label style={font=\Large},
  xlabel={Frame indexes},
  ylabel={PSNR},
  label style = {font=\LARGE}
]

\addplot[color=green!70!black, ultra thick, solid, mark=none]         
coordinates {\framewisePSNRARFree};
\addlegendentry{Ours}
\addplot[color=red, ultra thick, dashed, mark=none]         
coordinates {\framewisePSNRMCVD};
\addlegendentry{MCVD}
\addplot[color=blue, ultra thick, dotted, mark=none]         
coordinates {\framewisePSNRSTDiff};
\addlegendentry{STDiff}
\end{axis}
\end{tikzpicture}
\hspace*{5mm}

\begin{tikzpicture}%
\begin{axis}[
  width=100mm, height=70mm,
  xmin={10}, xmax={40}, xtick={10, 15, 20, 25, 30, 35, 40}, xticklabels={$10$, $15$, $20$, $25$, $30$, $35$, $40$},
  ymin={0}, ymax={0.3}, ytick={0.05, 0.1, 0.15, 0.2, 0.25}, yticklabels={$0.05$, $0.1$, $0.15$, $0.2$, $0.25$},
  grid={major}, legend pos={north east}, legend cell align={left}, legend style={font=\large},
  tick label style={font=\Large},
  xlabel={Frame indexes},
  ylabel={LPIPS},
  label style = {font=\LARGE}
]
\addplot[color=green!70!black, ultra thick, solid, mark=none]         
coordinates {\framewiseLPIPSARFree};
\addlegendentry{Ours}
\addplot[color=red, ultra thick, dashed, mark=none]                
coordinates {\framewiseLPIPSMCVD};
\addlegendentry{MCVD}
\addplot[color=blue, ultra thick, dotted, mark=none]                
coordinates {\framewiseLPIPSSTDiff};
\addlegendentry{STDiff}
\end{axis}
\end{tikzpicture}
}
\vspace{-1.5pc}
\caption{\label{fig:frame-wise comparison}%
Frame-wise comparisons with the KTH dataset using PSNR and LPIPS ($F_\text{total} = 30$).
}
\vspace{-1.2pc}
\end{figure}

\vspace{-0.7pc}
\subsection{Comparisons between different ARFree variants}
\vspace{-0.5pc}
We evaluated the two key components of ARFree: 
the proposed ARFree motion prediction module (A) in Section~\ref{sec: motion prediction} 
and the proposed training method (B) in Section~\ref{sec: training method}. 
Comparing the results in the first and second rows in Table~\ref{table:ablation} shows that the proposed motion prediction module leads to a performance improvement by modeling the temporal relation between the context and future frame tuples.
Comparing the results in the second and fourth rows in Table~\ref{table:ablation} demonstrates that the proposed training method enhances overall performance by improving motion continuity and contextual consistency between adjacent future frame tuples.

\vspace{-0.7pc}
\subsection{Frame-wise performance comparisons of long-term video prediction}
\vspace{-0.5pc}
Fig.~\ref{fig:frame-wise comparison} demonstrates that the proposed ARFree video prediction framework shows lower performance degradation in distant future frames compared to the existing SOTA methods.
This result imply that our framework can mitigate the error propagation issue in long-term video prediction.

\vspace{-0.7pc}
\section{Conclusion}
\vspace{-0.7pc}
Error propagation is a critical challenge in long-term video prediction.
To moderate this issue, we propose the first ARFree video prediction framework capable of predicting any future frame tuples, given the context frame tuple.
In future work, we aim to extend the proposed framework with datasets with more dynamic motions and improve its computational efficiency for real-time video prediction.

\newpage


\end{document}

%% file: data/framewise_results.tex
\newcommand{\framewisePSNRARFree}{%
(10, 32.98)
(11, 30.74)
(12, 29.42)
(13, 28.47)
(14, 27.75)
(15, 27.41)
(16, 26.95)
(17, 26.59)
(18, 26.23)
(19, 25.93)
(20, 25.77)
(21, 25.46)
(22, 25.20)
(23, 25.10)
(24, 24.90)
(25, 24.87)
(26, 24.74)
(27, 24.66)
(28, 24.55)
(29, 24.49)
(30, 24.38)
(31, 24.35)
(32, 24.23)
(33, 24.13)
(34, 23.98)
(35, 23.79)
(36, 23.72)
(37, 23.80)
(38, 23.85)
(39, 23.87)
}

\newcommand{\framewisePSNRMCVD}{%
(10, 35.87)
(11, 33.33)
(12, 31.55)
(13, 30.20)
(14, 29.31)
(15, 28.23)
(16, 27.77)
(17, 27.35)
(18, 26.96)
(19, 26.61)
(20, 26.00)
(21, 25.77)
(22, 25.53)
(23, 25.30)
(24, 25.08)
(25, 24.56)
(26, 24.40)
(27, 24.26)
(28, 24.12)
(29, 24.02)
(30, 23.69)
(31, 23.65)
(32, 23.58)
(33, 23.49)
(34, 23.41)
(35, 23.13)
(36, 23.16)
(37, 23.20)
(38, 23.28)
(39, 23.42)
}

\newcommand{\framewisePSNRSTDiff}{%
(10, 36.12)
(11, 31.94)
(12, 29.93)
(13, 28.48)
(14, 27.44)
(15, 26.67)
(16, 25.86)
(17, 25.24)
(18, 24.76)
(19, 24.30)
(20, 23.90)
(21, 23.57)
(22, 23.37)
(23, 23.16)
(24, 22.87)
(25, 22.64)
(26, 22.35)
(27, 22.11)
(28, 21.94)
(29, 21.78)
(30, 21.68)
(31, 21.59)
(32, 21.54)
(33, 21.49)
(34, 21.49)
(35, 21.49)
(36, 21.51)
(37, 21.62)
(38, 21.76)
(39, 21.93)
}

\newcommand{\framewiseLPIPSARFree}{%
(10, 0.0208)
(11, 0.0310)
(12, 0.0396)
(13, 0.0470)
(14, 0.0533)
(15, 0.0547)
(16, 0.0574)
(17, 0.0580)
(18, 0.0612)
(19, 0.0627)
(20, 0.0648)
(21, 0.0663)
(22, 0.0702)
(23, 0.0719)
(24, 0.0728)
(25, 0.0743)
(26, 0.0793)
(27, 0.0788)
(28, 0.0800)
(29, 0.0781)
(30, 0.0791)
(31, 0.0781)
(32, 0.0791)
(33, 0.0788)
(34, 0.0780)
(35, 0.0808)
(36, 0.0848)
(37, 0.0842)
(38, 0.0887)
(39, 0.0943)
}

\newcommand{\framewiseLPIPSMCVD}{%
(10, 0.0161)
(11, 0.0263)
(12, 0.0356)
(13, 0.0447)
(14, 0.0521)
(15, 0.0621)
(16, 0.0675)
(17, 0.0715)
(18, 0.0766)
(19, 0.0809)
(20, 0.0907)
(21, 0.0939)
(22, 0.0952)
(23, 0.0982)
(24, 0.1011)
(25, 0.1116)
(26, 0.1143)
(27, 0.1175)
(28, 0.1211)
(29, 0.1249)
(30, 0.1346)
(31, 0.1358)
(32, 0.1355)
(33, 0.1384)
(34, 0.13968)
(35, 0.1496)
(36, 0.1514)
(37, 0.1538)
(38, 0.1569)
(39, 0.1600)
}

\newcommand{\framewiseLPIPSSTDiff}{%
(10, 0.0109)
(11, 0.0235)
(12, 0.0339)
(13, 0.0433)
(14, 0.0512)
(15, 0.0579)
(16, 0.0651)
(17, 0.0710)
(18, 0.0745)
(19, 0.0785)
(20, 0.0858)
(21, 0.0894)
(22, 0.0931)
(23, 0.1005)
(24, 0.1063)
(25, 0.1104)
(26, 0.1179)
(27, 0.1226)
(28, 0.1262)
(29, 0.1331)
(30, 0.1388)
(31, 0.1431)
(32, 0.1451)
(33, 0.1494)
(34, 0.1526)
(35, 0.1594)
(36, 0.1639)
(37, 0.1701)
(38, 0.1745)
(39, 0.1781)
}